# Deep Ensembling for Perceptual Image Quality Assessment

Nisar Ahmed[1], H. M. Shahzad Asif[2,] Abdul Rauf Bhatti & Atif Khan

[1]Department of Computer Engineering, University of Engineering and Technology, Lahore, (54890) Pakistan.
[2]Department of Computer Science, University of Engineering and Technology, Lahore, (54890) Pakistan.
Corresponding author: Nisar Ahmed (e-mail: nisarahmedrana@yahoo.com).

**ABSTRACT:** Blind image quality assessment is a challenging task particularly due to unavailability of reference information. Training a deep neural network requires a large amount of training data which is not readily available for image quality. Transfer learning is usually opted to overcome this limitation and different deep architectures are used for this purpose as they learn features differently. After extensive experiments, we have designed a deep architecture containing two CNN architectures as its sub-units. Moreover, a self-collected image database BIQ2021 is proposed with 12,000 images having natural distortions. The self-collected database is subjectively scored and is used for model training and validation. It is demonstrated that synthetic distortion databases cannot provide generalization beyond the distortion types used in the database and they are not ideal candidates for general-purpose image quality assessment. Moreover, a largescale database of 18.75 million images with synthetic distortions is used to pre-train the model and then retrain it on benchmark databases for evaluation. Experiments are conducted on six benchmark databases three of which are synthetic distortion databases (LIVE, CSIQ & TID2013) and three are natural distortion databases (LIVE Challenge Database, CID2013 & KonIQ-10k). The proposed approach has provided a Pearson correlation coefficient of 0.8992, 0.8472 and 0.9452 subsequently and Spearman correlation coefficient of 0.8863, 0.8408 and 0.9421. Moreover, the performance is demonstrated using Perceptually Weighted Rank Correlation (PWRC) to indicate perceptual superiority of the proposed approach. Multiple experiments are conducted to validate the generalization performance of the proposed model by training on different subsets of the databases and validating on the test subset of BIQ2021 database.

# 1    Introduction

Image Quality Assessment (IQA) is a crucial and challenging task and required for many image processing applications [1]. IQA systems try to learn relationship between the image and its relative quality score provided by human observers in subjective quality assessment experiments. This quality score is a Mean Opinion Score (MOS) of human judgment. The relationship between image and its corresponding quality score is dependent on the human visual system which is a naively understood area and therefore modeling such a system is tough [2].

IQA has several applications as a quality metric [3-5]. It can be used to benchmark an image processing algorithm. If we have to select an image processing algorithm among several choices, a quality metric can help us to identify the one with best reproduced image quality. Similarly, it can be used to monitor image quality such as in a video transmission network; the metric can inspect the image/video quality and control the streaming. In a video or image acquisition system; a quality metrics can monitor and adjust the acquisition hardware parameters to obtain best quality image/video. Moreover, such quality metrics can be embedded into an image processing system for optimization of an algorithm's parameters to obtain best quality image e.g. image enhancement or visual communication systems. In short, IQA is crucial in numerous application scenarios where it can be used as benchmarking, optimization, control or monitoring setting.

The ultimate way to assess image quality is by visual inspection by humans [6]. It can be performed by subjective assessment and averaging the opinion of several subjects that is an expensive, cumbersome and difficult process. Objective quality assessment is sought for this scenario where an algorithm performs the task of IQA but it must correlate well with human judgment. The conventional image quality assessment algorithms such as PSNR or SSIM [7] are regarded as full-reference IQA methods. These methods require a reference image to perform quality assessment by learning some error/structural change to perform the task of quality assessment. These approaches have limited utility and are useful only when the algorithm has access to reference information. A more robust system which can model the behavior of human visual system and can work in no-reference setting is desired.

Classical no-reference IQA approaches extract natural scene statistics which are affected by degradations in an image and train a regression algorithm [8-10]. On the contrary, deep learning approaches especially Convolutional Neural Networks (CNN) have shown some promising performance in image quality assessment [11-13]. The success of CNN in IQA is due to the fact that it is inspired by the human visual cortex and it can learn quality-aware features by itself provided with representative training data. A problem in training a CNN for image quality assessment is lack of sufficient training data. This lack of sufficient data is tried to overcome by using transfer learning, data augmentation or training a relatively shallow CNN. Despite a lot of success in IQA by deep learning, there is still room for improvement as newer architectures and approaches provide better quality assessment performance and unveil methods that can result in true human visual system representation.

In this work, we have designed two, closely related, deep ensemble-based architectures. Contrary to our previous work [6] which uses multiple snapshots of training with a cyclic learning rate to construct an ensemble. The proposed deep ensemble can be regarded as a single model and trained end-to-end making it easier to train and it provide better quality assessment performance. The intuition of this architecture is based on idea that image quality depends on microstructures such as pixel relations and macro structures such as objects of interest. Different CNN architectures learn these features differently and their features can be combined for better representation. The proposed architecture, therefore, contains two entirely different classes of CNNs as a subset. The features of these two networks are concatenated and passed through global average pooling and few fully connected layers. Moreover, an image database BIQ2021 with natural distortions having 12,000 images is proposed. Extensive experiments are conducted to train the proposed DeepEns and DeepEns-Lite models on different subsets of databases and perform validation on the test set of the proposed database BIQ2021. The objective of the experimentations is to demonstrate the usefulness of the proposed end-to-end training approach for general-purpose image quality assessment. The proposed image database is used for a similar purpose to highlight that model trained on image databases with simulated distortions does not perform well on images with natural distortions. Specific contributions of this work are highlighted below:

1. A large-scale image database (12,000 images) having natural distortions and laboratory-controlled subjective scores is introduced.
2. Deep ensemble-based architectures are proposed for end-to-end training.
3. A deep CNN training approach is proposed using synthetic distortion database with 18.75 million images.
4. End-to-end training of deep CNN architecture and comparison with existing approaches to demonstrate effectiveness.
5. Cross-validation of the trained model on natural distortion databases to highlight its generalizability.
6. Demonstration of the fact that image databases with simulated distortions are not suitable for training of models for general-purpose image quality assessment.

## 2 Related Work

This section provides a brief overview of recent work which is closely related to the proposed approach. The problem of blind image quality assessment is conventionally addressed based on Natural Scene Statistics (NSS) [1]. The surprising performance of deep learning in visual recognition has directed the IQA research from NSS to deep learning. An important advantage of deep learning-based methods is that there is no need for handcrafted features as they are learned directly based on training images. The difficulty in deep learning-based methods lies in availability of training data as the available training image for IQA are not very large. However, researchers have developed techniques that can address the problem of IQA using deep learning-based methods.

Kang et al. [14, 15] proposed a deep model which trains the CNN using spatially normalized image patches. The quality and distortion types are identified simultaneously using a multi-task CNN architecture. Bianco et al. [12] have proposed DeepBIQ which is a blind IQA model based on CNN. They have used an AlexNet like architecture which is pretrained on ImageNet. They have extracted features from this deep network by taking multiple crops of 224x224 and then average-pooled the features to train a Support Vector Regression (SVR). Gao et al. [16] have used a Vgg16 architecture for quality assessment and reasoned that different levels of convolution represent image quality differently. They trained an SVR on features extracted at different levels of convolution and then averaged pooled to provide image quality. Kim et al. [17] proposed a two-stage image quality predictor. The first stage predicts an objective error map and the second stage predicts subjective quality score. Ravela et al. [18] has also opted for a two-stage approach. The first stage predicts the distortion type and the second stage uses a specialized deep model to predict the subjective score and perform average pooling with the prediction of other deep models. Ma et al. [19] also proposed a two-stage network: the first identifies the type of distortion and the second stage performs the subjective score prediction using a specialized quality prediction network for each distortion type.

Zhang et al. [20] also proposed a two-stage framework, the first stage identifies the distortion type as well as the level of distortion. The second stage performs quality assessment using a specialized CNN model. Their models have focused to perform quality-assessment for both authentically distorted as well as inauthentically distorted images. Fan et al. [21] also follow a two-stage approach by first predicting the distortion type and then performing subjective score prediction using multiple CNNs.

Deep features have provided unreasonable effectiveness for image quality assessment [22]. The problem with such methods is that they cannot be trained end-to-end but they are simple and effective. The features extracted from ImageNet pre-trained models such as VGG are used to train a regression algorithm for image quality assessment [11]. Ahmed [23] has experimented with feature extraction from different layers of ImageNet pre-trained models. Moreover, they have retrained these models on image quality databases and then performed the feature extraction. It is demonstrated that ImageNet pretrained models are not a good candidate for perceptual quality assessment of digital images. Retraining of these models makes them learn quality-aware features and these features can be used for image quality assessment. Another approach is to train the regression algorithm with natural scene statistics as well as deep-features for an enriched feature experience [2].

The proposed approach, on the other hand, follows a slightly different approach and proposes an architecture that uses two entirely different CNN architectures as its subunits. The EfficientNet-B0 [24] is a lightweight CNN model which uses inverted bottleneck units for its construction followed by pooling, fully connected and classification layers. NASNet-mobile [25] on the other hand uses a cell structure that is learned using neural architecture search on CIFAR-10 dataset. Both of them learn different types of visual features, are lightweight and provided superior performance to single computationally expensive CNN model. A novel training strategy is proposed for quality assessment tasks using synthetically distorted images. The proposed architecture is, therefore, more suitable at learning rich-feature set and in turn, provides quality scores which are comparable to the state-of-the-art.

## 3 Materials and Methods

IQA is a challenging task and numerous datasets have been published for full-reference as well as no-reference quality assessment experiments. We have selected six popular databases to perform experiments and benchmark the results. LIVE and TID are image quality databases released in three and two versions subsequently whereas CSIQ, LIVE in the wild Challenge Database (LiveCD), CID2013 and KonIQ-10k are released in a single version. Table 1 provides the eight versions of six databases with number of images, scoring method and range. We have linearly transformed the range of all the databases from 0 to 1 so they can be used for cross-dataset evaluation. It is highlighted that CSIQ, LIVE-I, LIVE-II, TID2008 and TID2013 are datasets generated from reference images by simulating distortions and are called synthetic distortion databases. LiveCD, CID2013 and KonIQ-10k are different types of dataset with images having naturally occurring distortions. LiveCD contains images that are captured by random devices and have some sort of distortions whereas CID2013 has used a fixed set of image acquisition devices with a set of parameters to introduce distortions intentionally. They have captured images in 36 different indoor and outdoor locations with different cameras and settings to capture the distortions occurring naturally. KonIQ-10k is the largest database introduced recently. It contains 10073 images collected from online sources and subjectively scored using crowdsourcing. Our proposed dataset contains all three types of images having natural distortions. The details of this database are described further in the next section.

**Table 1** Benchmark datasets with number of images, scoring method and range

| Dataset Name | Number of reference images | Number of distorted images | Scoring Method | Range |
|---|---|---|---|---|
| CSIQ [26] | 30 | 900 | DMOS | 0-1 |
| LIVE-I [27] | 29 | 460 | DMOS | 0-100 |
| LIVE-II [27] | 29 | 982 | DMOS | 0-100 |
| TID2008 [28] | 25 | 1700 | MOS | 1-10 |
| TID2013 [29] | 25 | 3000 | MOS | 1-10 |
| CID2013 [30] | Nil | 474 | MOS | 0-100 |
| LiveCD [31] | Nil | 1169 | MOS | 0-100 |
| KonIQ-10k | Nil | 10073 | MOS | 1-5 |
| BIQ2021 | Nil | 12000 | MOS | 0-1 |

### 3.1 BIQ2021

BIQ202 is the proposed database which contains images having natural distortions. It is a large-scale database of 12,000 images. The collection of the images for the dataset is performed in three different subsets: (i) The first subset contains 2,000 images which are captured with different camera settings (i.e. ISO, shutter speed, focal length and motion) to introduce distortions. This subset contains images captured from the same scene having degradations ranging from just noticeable to severe as depicted in Fig. 1. These images contain lack of focus, motion blur, non-uniform illumination, sensor noise and a mix of different degradations. (ii) The second subset contains 2,000 images with natural distortions introduced during the process of acquisition or storage. These images are not captured with intentions to be used for image quality research and are author's collection. These images, apart from the first subset contain compression, processing and storage-related degradations. Fig. 2 provides a subset of images for demonstration. (iii) The third subset contains 8,000 images that are manually selected from Unsplash.com having varying image quality and content. These images contain all the distortions occurring in the earlier subsets and they may involve degradations resulting due to pot-processing and trends of the photographic community. These images are added to introduce content diversity by downloading images having different tags such as animals, people, babies, sports, architecture, nature, etc. Moreover, this subset provides a representation of images captured by a typical photography community. Fig. 3 provides a depiction of the subset of images in this category.

Subjective scoring of these images is performed in laboratory settings during multiple sessions. Each session consisted of half to one hour. The session time, viewing distance and viewing angle were at the liberty of the subject so the quality scores can be obtained naturally. The ambient conditions, display device and scoring mechanism are kept constant. The display device used during the subjective scoring is HP 24 inches LED, full-HD display, 16:9 aspect ratio and 200 cd/m$^2$ brightness. The scoring is done through absolute category rating with five quality levels. The observer was free to use a continuous value through a slider or select a discrete rating. Fig. 4 provides the GUI of the desktop application used for subjective scoring. Moreover, the score history was provided to encourage the observer to use the full-range of scoring scale. The observer was free to leave the experiment if he feels withdrawn. Most of the observers were graduate or undergraduate students of computer science and engineering and had no specific expertise in the domain of image quality assessment. The experiments were terminated after 30 observers completed the subjective scoring. MOS is calculated by averaging the quality scores of 30 observers and then scaled to a range of 0-1. The variance of scores provided by 30 observers along with MOS are provided in the dataset. Fig. 5 provides the histogram of MOS with 100 bins.

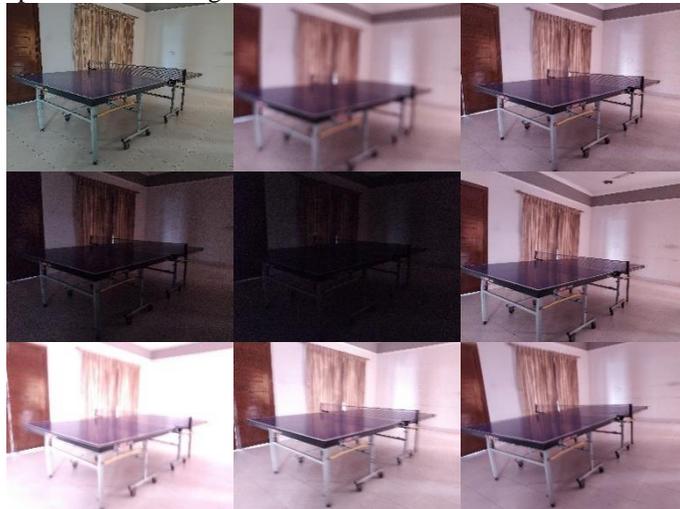

**Fig. 1**  Images captured with different camera settings

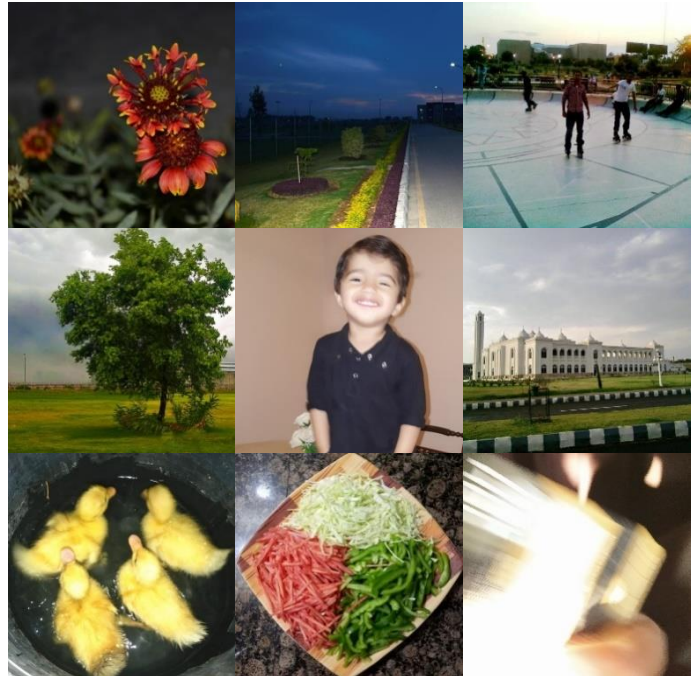

**Fig. 2** Images selected from random image collections

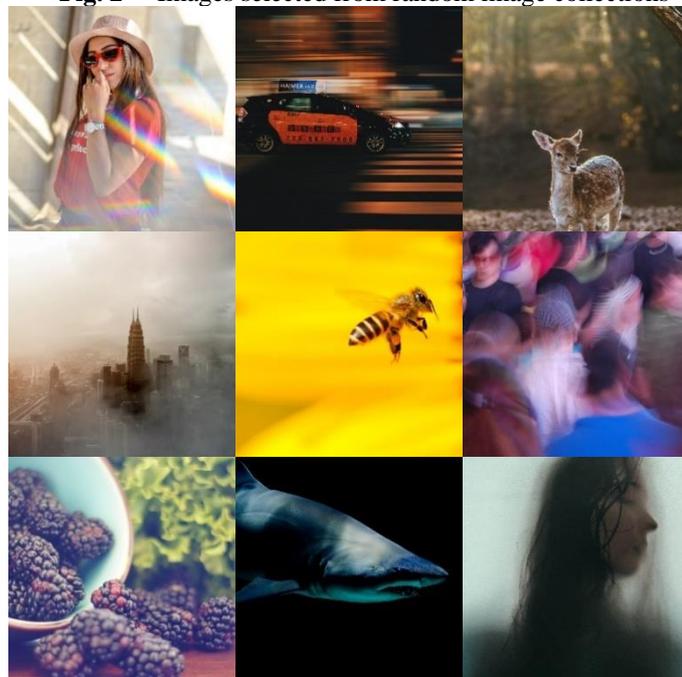

**Fig. 3** Images with diverse content downloaded from Unsplash.com

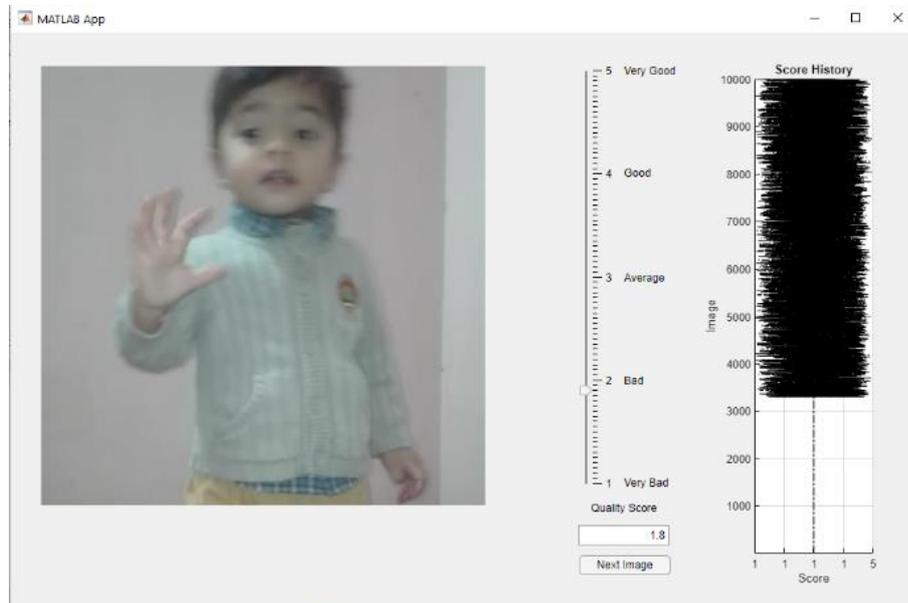

**Fig. 4** GUI used for subjective quality scoring

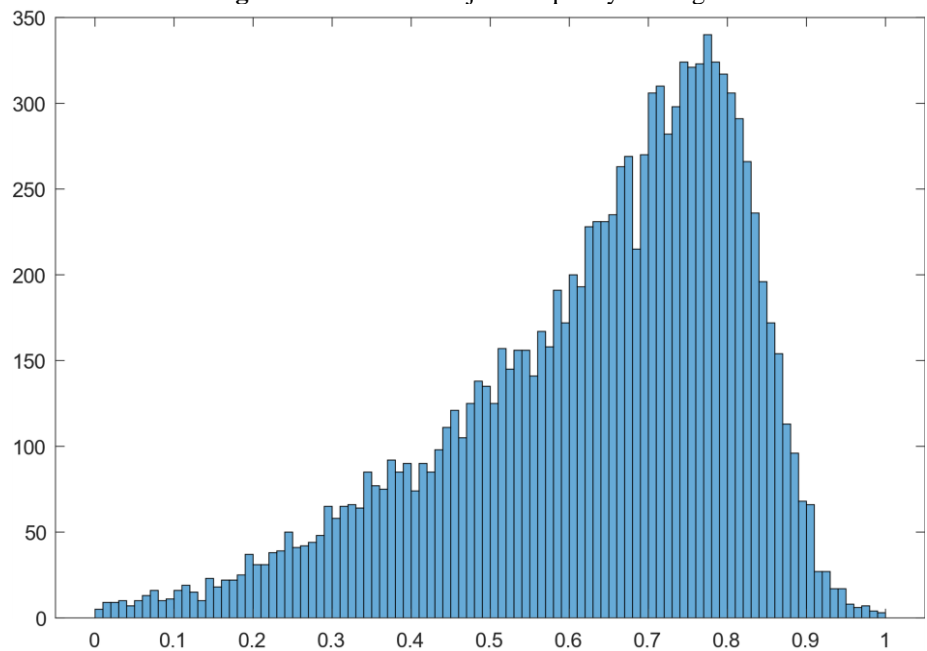

**Fig. 5** MOS distribution of BIQ2021

### 3.2 Deep Ensemble

CNNs are a class of neural networks designed specifically for visual recognition tasks. There are several pretrained models which are originally trained on ImageNet visual recognition challenge to classify images into 1,000 categories. Image quality assessment researchers either use these models for feature extraction or perform transfer learning for image quality assessment. These pretrained models have different representational power and learn different types of deep features. Fig. 6 and 7 provide DeepDream visualization of last fully connected layer for six channels (114, 293, 341, 484, 563 & 950) of two of these architectures with the same initialization. It can be seen that these features are visually different from each other. Moreover, it is demonstrated experimentally that they provide different quality assessment performance [23].

We have provided two versions of the proposed DeepEns architecture. The lighter version named DeepEns-Lite contains EfficientNet-B0 and NASNet-mobile as base architectures as they are small and have good visual recognition performance. The selection of these two base architectures is based on their lighter weight and good quality assessment performance. The full-version named DeepEns contains InceptionResNet-V2 and EfficientNet-B7 as base architectures as they have provided the highest quality assessment performances among 19 popular CNN models. The architectures of DeepEns and DeepEns-Lite are provided in Fig. 8 & 9. We have repurposed these models for our problem and taken the CNN base to construct our architecture which extracts features by both and then concatenates them. Although there are different global pooling operations max-pooling and average-pooling are the most common ones. Guo et al. [32] have presented a study to analyze the effect of nine pooling strategies on fine-grained visual recognition. They have concluded that max-pooling learn the discriminative details by learning smaller and important distinguishing parts of the image whereas the average-pooling provide an averaging effect and learn global image parameter and doesn't learn specific details. As the image quality is a global parameter and is not specific to shapes or areas, we have therefore used global average-pooling. Although a thorough study can be conducted to study the effect of different pooling strategies on image quality assessment but it is not done here and left as future work.

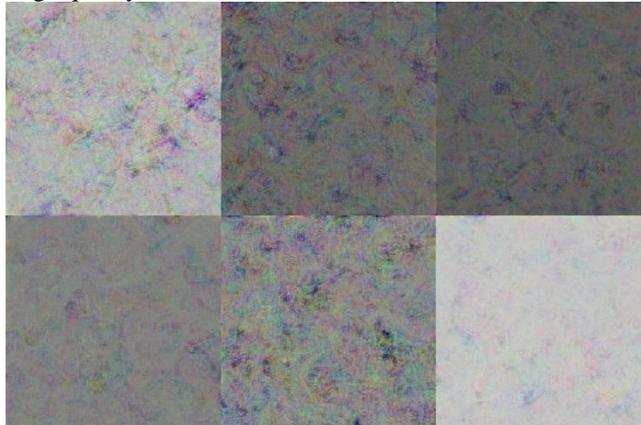

**Fig. 6** Deepdream visualization of features learned by NASNet-Mobile

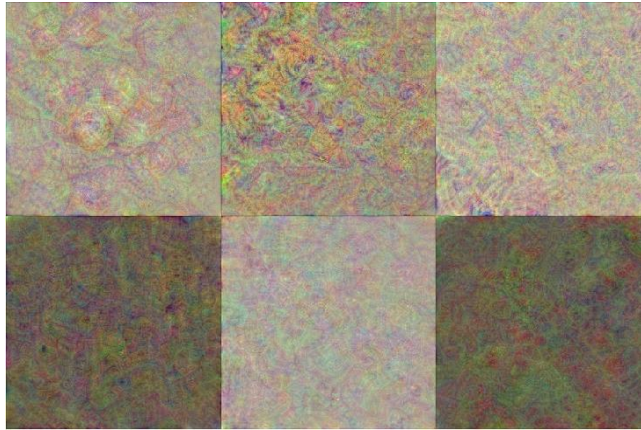

**Fig. 7** Deepdream visualization of features learned by EfficientNet-B0

Three fully connected layers with 1024, 256 and 1 neurons preceded the global average-pooling layer are used in lighter version DeepEns-Lite. In the full-version, global average-pooling is applied on both base architectures prior to concatenation. The concatenation is followed by 4096, 1024, 256 and 1 neurons. The dropout 25% is used in earlier layers and 50% in the layer with 256 neurons which provides additional regularization. The Rectified Linear Unit (ReLU) is used as activation function with all the fully-connected layers. The final fully connected layer contains one neuron and provides the activations to regression layer.

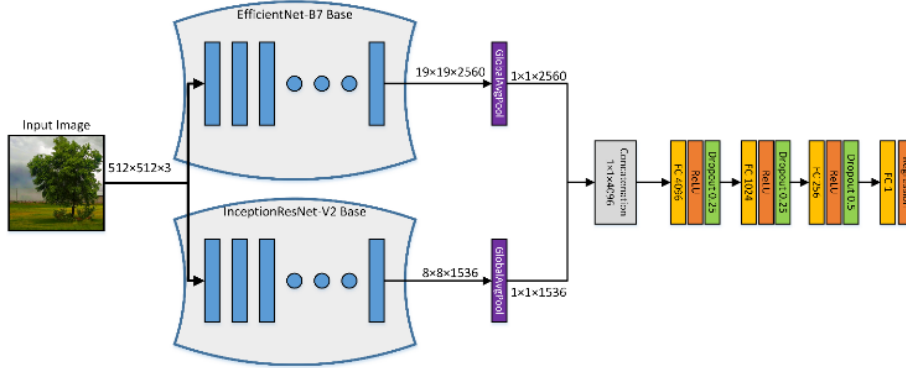

Fig. 8 Architecture of DeepEns

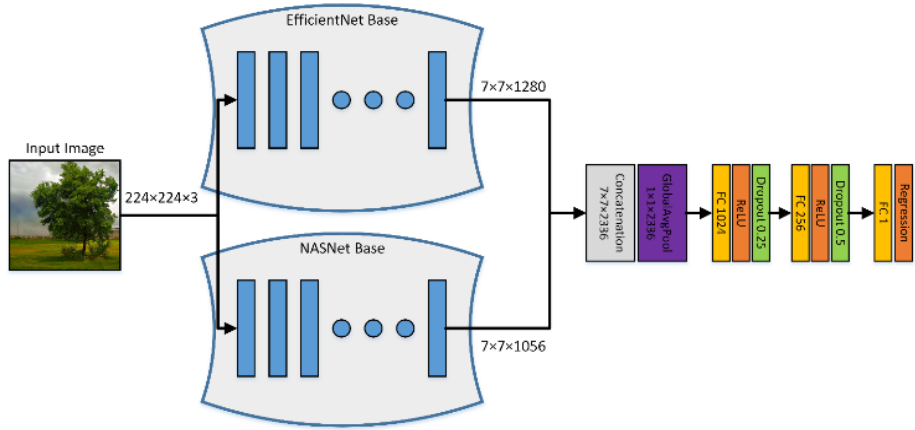

Fig. 9 Architecture of DeepEns-Lite

### 3.3 Loss Function

The regression layer normally uses mean squared error (MSE) as a loss function for training. The advantage of MSE is that it provides a smooth convex function which is easier to train due to ease in gradient computation. Although, mean squared error seems fine as a loss function but keeping in view the trend in literature we have experimented with Mean Absolute Error (MAE), Mean Absolute Percentage Error (MAPE), Mean Squared Logarithmic Error (MSLE), LogCosh loss and Huber loss along with MSE. It is observed that mean squared error provides faster training and the performance obtained with model trained using MSE are slightly better than the others. Table 1 provides the validation RMSE-values obtained after training the DeepEns-Lite for 30 epochs. The experiments are conducted using LiveCD. The details of these loss functions along with their training progress are provided in Appendix A. It is evident from the training progress of Fig. A-1 to Fig. A-6 and Table 2 that MSE is the best candidate loss function among the six loss functions. The better performance of MSE is possibly due to the fact that MOS is obtained through controlled experiments with outlier rejection and is average of a number of observers. It is therefore less possible to experience outliers and other abnormalities in training data and therefore quadratic nature of MSE make it a better candidate. Although, visual perception-based loss functions, such as the one based on visual saliency may perform better but they are not explored in this study and left as a future work.

**Table 2** Loss function performance for 30 epochs on LiveCD database

| Sr. | Loss Function | RMSE |
| --- | --- | --- |
| 1 | Mean Squared Error | 1.1403 |
| 2 | Mean Absolute Error | 1.9873 |
| 3 | Mean Absolute Percentage Error | 2.1201 |
| 4 | Mean Squared Logarithmic Error | 1.5501 |
| 5 | LogCosh Loss | 1.2553 |
| 6 | Huber Loss | 1.9315 |

### 3.4 Model Pretraining

Pretraining of deep CNN models is required as the benchmark database sizes are not sufficient to train a model from scratch. The largest benchmark database has 10,000 images. It is the practice in the literature to use models pretrained on ImageNet and repurpose them to be used for image quality assessment through transfer learning. The authors have used a different strategy to train the models for quality assessment apart from ImageNet. We have used 150,000 images provided by Kadis-700K [33] to generate distorted images. We have generated 25 different types of distortions in 5 distinct levels and obtained 18.75 million distorted images and named Kadid-19M. We have assigned distortion type and level as the label of each image such as '23_3' for distortion number 23 with distortion level of 3. This way, we had a labeled database of 18.75 million images with 125 categories. This is a large database and is different from ImageNet. The models trained on ImageNet are specialized for visual recognition and can be transferred to other related tasks. Image quality assessment is an entirely different problem as images with entirely different content may have same quality score or images of same scene may have different quality scores. It is therefore proposed to use a database which is related although having less reliable labels. This proposed strategy is found to be more useful in performing transfer learning. The performance of the proposed strategy is demonstrated using EfficientNet-B0 on TID2013 image database in Table 3. The training was performed with a piecewise learning rate until convergence.

Table 3 Transfer Learning for image quality

| Metric | ImageNet | Kadid-19M |
|--------|----------|-----------|
| RMSE   | 0.4561   | 0.4426    |
| PLCC   | 0.9471   | 0.9490    |
| SROCC  | 0.9427   | 0.9488    |

### 3.5 Training Strategy

Same training options are used for training of pretrained model and the transfer learning only difference was the choice of learning rate and epochs. The initial training using synthetic database used adam optimizer with an initial learning rate of $1e^{-2}$ with a piecewise learning rate with a drop factor of 50% and a drop period of 10 epochs. The training was performed for 100 epochs with a batch size of 16.

Regularizations are important to improve the generalization of deep CNN models. We have incorporated dropout in the fully connected layers to reduce the overfitting. Similarly, we have incorporated data augmentation to reduce overfitting and improve generalization. We have used two image augmentation strategies namely random horizontal reflection and random rotation. Some image augmentation strategies such as scaling, shear, contrast, equalization, color and brightness variation are not used as the perceptual quality is sensitive to these augmentations. Random cropping also serves the purpose of regularization and therefore horizontal or vertical translation is not necessary.

As the input size of the proposed architecture is $224 \times 224$ incase of DeepEns-Lite and $512 \times 512$ in case of DeepEns but the image size in the different databases is different. One method to match the input resolution is to resize but it affects the perceptual quality of the image as demonstrated by [34]. The second approach is center cropping which doesn't provide a regularization effect and may be less useful in case of non-uniform illumination, narrow depth of field and varying information in different image regions. The third approach is random cropping in which a randomly selected image patch equal to input resolution is cropped in each epoch and therefore resolves the issue of input size inconsistency as well as provides regularization. This will also provide averaging effect over different epochs and will be a close representation of the image. The training progress of DeepEns-Lite is provided in Fig. 10 for demonstration.

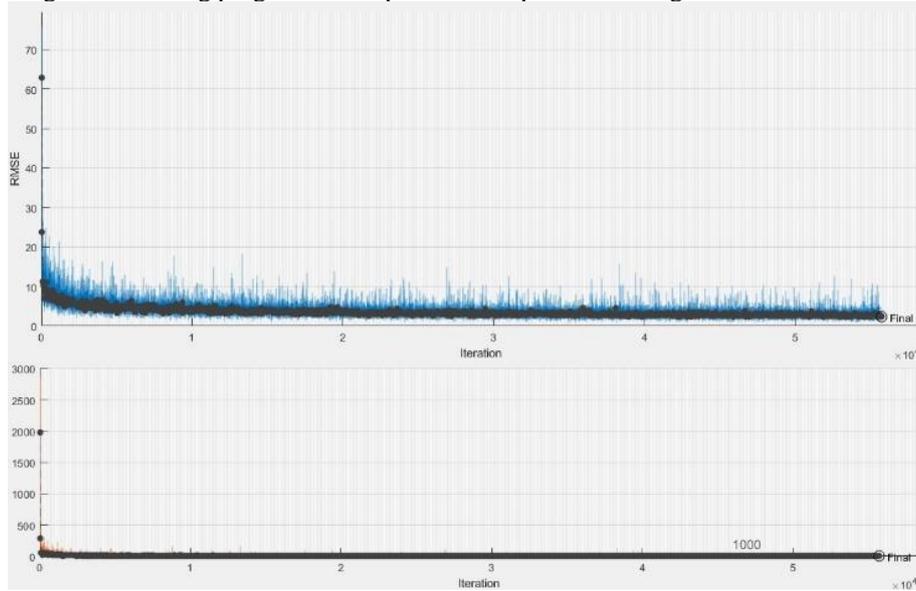

**Fig. 10** Training Progress of the DeepEns-Lite architecture

In the testing phase, the trained network performs predictions on N randomly cropped region of each image and their scores are averaged. Given an input image divided into $N_c$ random crops and each having a quality score $q_i$. The predicted quality score is calculated by averaging the individual quality score of each crop by using Equation (1).

$$q = \frac{1}{N_c} \sum_i^{N_c} q_i \ldots (1)$$

## 4 Experiments

This section describes the experimental setup used for evaluation of the proposed architectures. The experiments are conducted by taking $224 \times 224$ image patches in case of DeepEns-Lite and $512 \times 512$ in case of DeepEns. The predicted quality scores reported in this sections are average of 10 crops calculated using formula in equation 1. The experiments are performed on Dell T5600 workstation with two Intel E5-2687W CPU, 32GB RAM and RTX 2070 GPU. The dataset is stored on SATA SSD to reduce the preprocessing and mini-batch loading latency.

### 4.1 Evaluation Metrics

The performance of the proposed approach is assessed using four different metrics which measures correlation or error between subjective and predicted quality scores.

Root Mean Squared Error (RMSE) measures the error between the subjective score $y_S$ and the predicted score $y_P$. It measures how accurately the model has predicted the quality score. A smaller value of RMSE means the model provides predictions with less average deviation from subjective scores. Equation (2) provides the formula for RMSE calculation.

$$RMSE = \sqrt{\frac{\sum_i (y_P - y_S)^2}{n-1}} \dots (2)$$

Pearson Linear Correlation Coefficient (PLCC) measures the linear relation between subjective score $y_S$ and the predicted score $y_P$. A higher value of PLCC indicates that predicted scores are consistent with human judgment. Equation (3) provides the formula for PLCC calculation.

$$PLCC = 1 - \frac{\sum_i (y_P - y_S)(\hat{y}_P - \hat{y}_S)}{\sqrt{\sum_i (y_P - y_S)^2}\sqrt{(\hat{y}_P - \hat{y}_S)^2}} \dots (3)$$

Spearman Rank Order Correlation Coefficient (SROCC) measures the correlation with ranked subjective scores. A higher value of SROCC indicates that the trend of subjective score is correlated with that of predicted score which indicates consistency with human judgment. Equation (4) provides the formula for SROCC computation.

$$SROCC = 1 - \frac{6 \sum_n d_n^2}{N(N^2 - 1)} \dots (4)$$

Perceptually Weighted Rank Correlation (PWRC) [35] is a perceptual approach for comparison of objective image quality assessment algorithms. It works by rewarding the capability of correct ranking of high-quality images and suppressing the attention towards insensitive rank mistakes. The PWRC calculates the area under the curve for $S(x, y, T)$ which is a combination function that fuses three rank correlation components. The components of the combination function are perceptually weighted activation function, outlier detection function and importance measurement function.

$$PWRC = \int_{T_{min}}^{T_{max}} S(x, y, T) dT$$

where

$$S(x, y, T) = f[A(x, T), D(p, q), M(p, q)]$$

### 4.2 Performance Assessment on Individual Datasets

The performance assessment test is performed on the 20% hold-out dataset which is not exposed during training. The model is trained on 80% dataset and the testing results are evaluated using three performance metrics described earlier. Table 4 provides the results of experimental evaluation based on defined metrics on six datasets. It is worth noting that LIVE is combination of LIVE release-1 and LIVE release-2 as they have followed the same scoring methods but different distortion categories. TID2008 and TID2013 cannot be combined this way as they have overlapping distortions and are independently reported in the literature. It can be noted from Table 4 that the synthetically distorted image databases can be easily modeled due to limited number of distortions simulated on a set of reference images. The naturally distorted image databases are difficult to model as they contain distortions introduced during the process of image acquisition either intentionally or unintentionally.

**Table 4** Experimental evaluation on individual datasets

|       | LIVE   | TID2013 | CSIQ    | LiveCD  | CID2013 | KonIQ-10K |
|-------|--------|---------|---------|---------|---------|-----------|
|       |        |         | **DeepEns-Lite** |  |    |           |
| RMES  | 0.0825 | 0.4212  | 0.1117  | 0.1095  | 0.1414  | 0.1011    |
| PLCC  | 0.9813 | 0.9738  | 0.9784  | 0.8992  | 0.8472  | 0.9452    |
| SROCC | 0.9781 | 0.9761  | 0.9748  | 0.8863  | 0.8408  | 0.9421    |
| PWRC  | 8.2121 | 7.9931  | 10.2141 | 11.0214 | 13.7723 | 12.1641   |
|       |        |         | **DeepEns** |     |         |           |
| RMES  | 0.0569 | 0.4170  | 0.1074  | 0.0834  | 0.1268  | 0.0859    |
| PLCC  | 0.9860 | 0.9874  | 0.9846  | 0.9135  | 0.8620  | 0.9478    |
| SROCC | 0.9887 | 0.9818  | 0.9817  | 0.8985  | 0.8421  | 0.9442    |
| PWRC  | 8.2136 | 8.3944  | 10.2157 | 12.0242 | 13.7752 | 14.1769   |

It is worth mentioning, as the benchmark image databases don't have a train/test split, it is therefore not possible to make a fair comparison. We have therefore followed the literature in which 80% of data is randomly selected for training and 20% for testing. The experiment is repeated several times to make the performance independent of the split as much as possible. We have repeated this experiment 10 times as the training of a CNN is a computationally expensive task and repeating it for larger number of times is not feasible for us. The histograms of three of these performance parameters are provided for CSIQ and CID2013 in Fig. 11 & 12 respectively. It can be noted from the Fig.s that the performance metrics are close to median value and follow nearly a normal distribution.

The scatter plot of ground-truth vs predicted values along with regression line fitting is provided in Fig. 13. The scatter plots are provided for the train-test split with median value of PLCC. In CSIQ database Fig. 13 (a), 175 samples are used for testing with most of the samples very closely predicted to their ground-truth scores. In CID2013 database Fig. 13 (b) 93 samples are used for testing with samples evenly distributed along the regression line.

The residuals play an important role in analysis of a regression model. We have chosen CSIQ image database to perform the residual analysis whereas similar results were obtained using other databases. The bar-plot as depicted in Fig. 14 (a) is a useful method to visualize the magnitude and direction of the residuals and can be used for small to medium number of samples. The box-plot can provide the mean and the distribution of residual values in different iterations. Fig. 14 (b) provides the box-plot for 25 iterations drawn randomly for better visualization. It can be seen that the variation in the mean value among different iterations doesn't vary largely and the absolute magnitude goes up to 1.5 which is even less in positive direction. The scatter-plot between ground-truth on horizontal axes and residual magnitude on vertical axes is a very important graph that help in understanding the model's behavior. A skewness in this distribution such as the points following a certain shape such as cone or parabola or skewed to lower or the upper extreme indicates some limitations in the modeling process. A random distribution that doesn't follow any clear pattern indicates the model is reasonable and the same is indicated in Fig. 14 (c). Another important measure in residual analysis is that the residuals should follow a normal distribution. We have plotted a probability plot for normal distribution on the residuals in Fig. 14 (d) and it can be observed that most of the points fall on the line except for very few deviations. It indicates that the residuals follow a normal distribution for the proposed model.

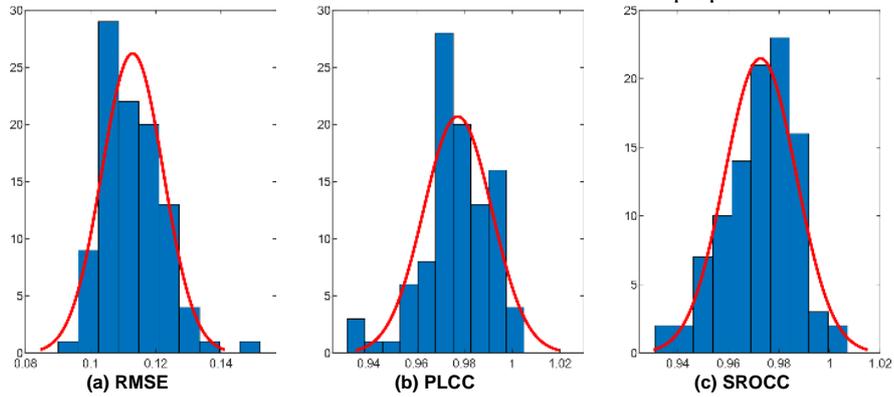

**Fig. 11**  CSIQ: Histograms of (a) RMSE, (b) PLCC & (c) SROCC for 100 random train-test splits

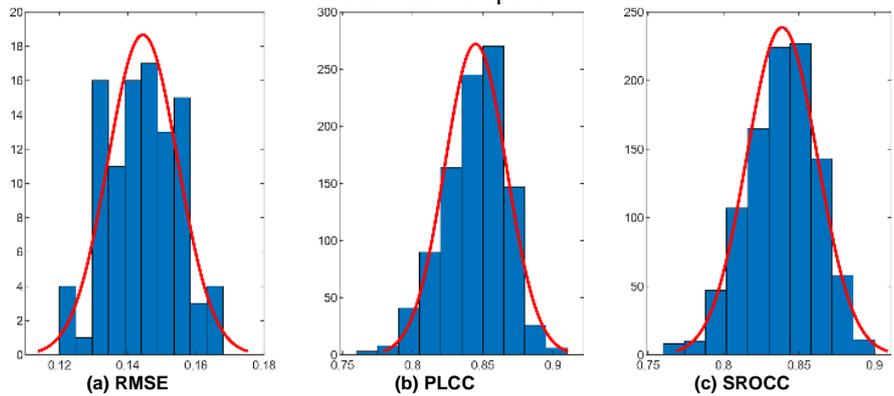

**Fig. 12**  CID2013: Histograms of (a) RMSE, (b) PLCC & (c) SROCC for 100 random train-test splits

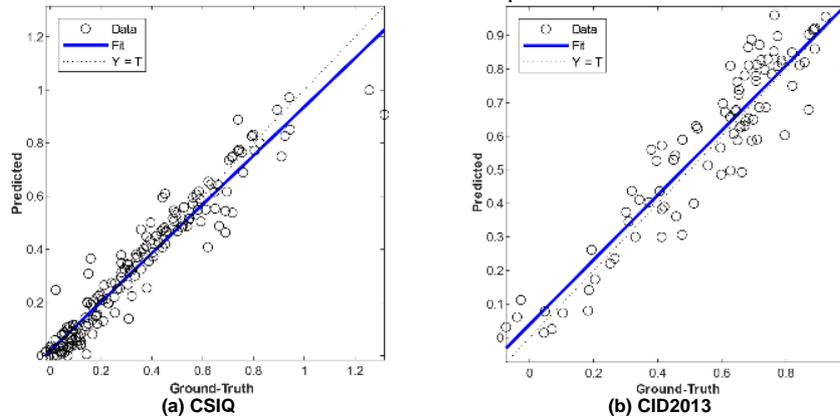

**Fig. 13**  Scatter plot between ground-truth vs predicted values along with regression line for (a) CSIQ database & (b) CID2013 database

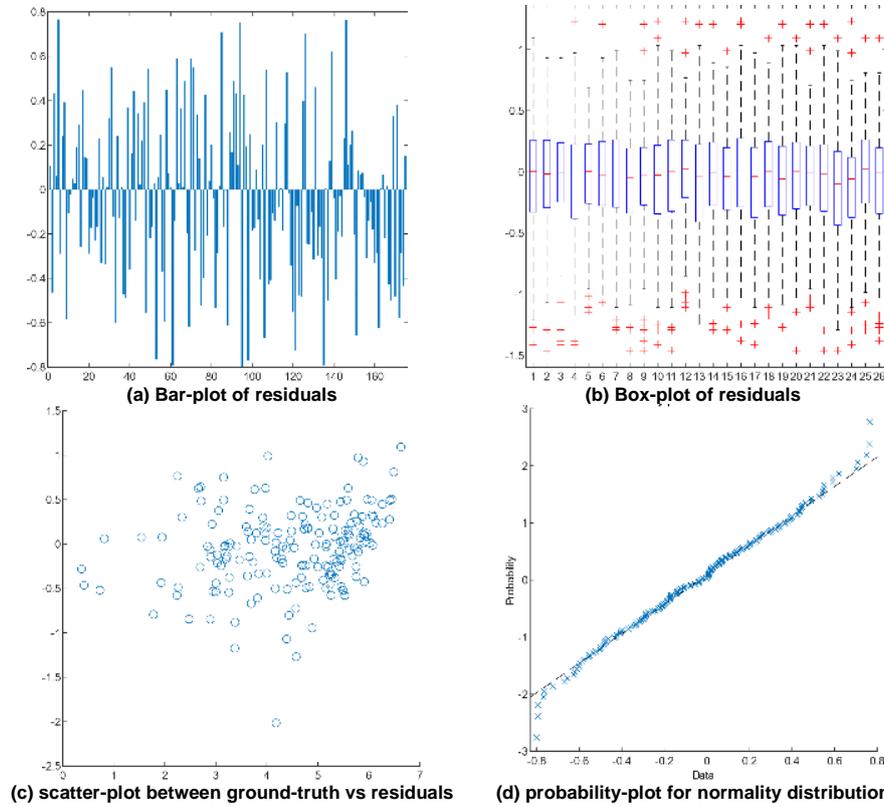

**Fig. 14** (a) CSIQ Database: Bar-plot of residuals, (b) box-plot of residuals for 25 iterations, (c) scatter-plot between ground-truth vs residuals and probability-plot for normal distribution.

### 4.3 Comparison with Existing Approaches

This section provides a comparison with the existing approaches. We have performed comparison with seventeen existing approaches and some of them are discussed in related work due to similarity of their methodology with the DeepEns. It is worth mentioning that most of the top-performing approaches are based on deep learning and their contribution lies in architecture or training methodology. Moreover, we have performed comparison on four benchmark datasets but not all the techniques have reported results on these datasets. Moreover, the reported results are as per respective author's claim and they have not been computed again due to constraints of time and availability of source code. The reported score is provided for up to four decimal places for the DeepEns models but some of the approaches have reported scores up to two decimal places and they are mentioned as it is. RMSE is not reported by most of the approaches, therefore it is eliminated from the comparison and only PLCC and SROCC are mentioned in Table 6. Scores for TID2008 are reported by only Kang et al. [15] with values of 0.903 for PLCC and 0.999 for SROCC and therefore not mentioned in the table. It can be seen that DeepEns has provided highest performance in all the categories whereas DeepBIQ [12] stands as second best approach. The results for PWRC and comparison of existing approaches is provided in Table 5 and the scores of existing approaches are taken from [35].

**Table 5** Comparison of the DeepEns with existing approaches

|              | LIVE   |        | TID2013 |        | LiveCD |         |
|--------------|--------|--------|---------|--------|--------|---------|
|              | SROCC  | PWRC   | SROCC   | PWRC   | SROCC  | PWRC    |
| BIQI         | 0.784  | 4.195  | 0.674   | 3.701  | 0.522  | 9.193   |
| BLIINDS II   | 0.925  | 5.362  | 0.839   | 5.04   | 0.489  | 8.593   |
| BRISQUE      | 0.925  | 5.374  | 0.865   | 5.303  | 0.592  | 10.55   |
| DIIVINE      | 0.9    | 5.17   | 0.775   | 4.495  | 0.576  | 10.413  |
| NFERM        | 0.933  | 5.472  | 0.888   | 5.509  | 0.579  | 10.165  |
| M3           | 0.951  | 5.659  | 0.89    | 5.491  | 0.598  | 10.479  |
| TCLT         | 0.942  | 5.642  | 0.932   | 5.964  | 0.561  | 9.601   |
| DeepEns-Lite | 0.9781 | 6.8121 | 0.9761  | 6.4931 | 0.8863 | 11.0214 |
| DeepEns      | 0.9887 | 8.2136 | 0.9818  | 8.3944 | 0.8985 | 12.0242 |

**Table 6** Comparison of the DeepEns with existing approaches

| Sr. | Dataset           | LIVE   |        | TID2013 |        | CSIQ   |        |
|-----|-------------------|--------|--------|---------|--------|--------|--------|
|     | Metric »→▶        | PLCC   | SROCC  | PLCC    | SROCC  | PLCC   | SROCC  |
| 1   | Kang et al. [14]  | 0.953  | 0.956  | -       | -      | -      | -      |
| 2   | DeepBIQ [12]      | 0.98   | 0.97   | 0.96    | 0.96   | 0.97   | 0.96   |
| 3   | BLINDER [16]      | 0.959  | 0.966  | 0.838   | 0.819  | 0.968  | 0.961  |
| 4   | DIQA [17]         | 0.977  | 0.975  | 0.850   | 0.825  | 0.915  | 0.884  |
| 5   | Ravela et al. [18]| 0.9492 | 0.9492 | -       | -      | 0.9445 | 0.9445 |
| 6   | Meon [19]         | -      | -      | 0.912   | 0.912  | 0.944  | 0.932  |
| 7   | DB-CNN [20]       | 0.971  | 0.968  | 0.865   | 0.816  | 0.959  | 0.946  |
| 8   | MCNN-IQA [21]     | 0.957  | 0.9531 | -       | -      | 0.894  | 0.8766 |
| 9   | DIQaM-NR [36]     | 0.972  | 0.960  | 0.855   | 0.835  | -      | -      |
| 10  | WaDIQaM-NR [36]   | 0.963  | 0.954  | 0.787   | 0.761  | -      | -      |
| 11  | OG-IQA [37]       | 0.952  | 0.950  | -       | -      | -      | -      |
| 12  | VIDGIQA [38]      | 0.973  | 0.969  | -       | -      | -      | -      |
| 13  | Bosse et al. [39] | 0.972  | 0.960  | -       | -      | -      | -      |
| 14  | Bare et al. [40]  | 0.974  | 0.971  | -       | -      | -      | -      |
| 15  | dipIQ [41]        | 0.958  | 0.957  | 0.894   | 0.877  | 0.949  | 0.93   |
| 16  | HOSA [42]         | 0.950  | 0.952  | 0.952   | 0.959  | 0.930  | 0.948  |
| 17  | BIECON [43]       | 0.962  | 0.958  | 0.765   | 0.721  | 0.838  | 0.825  |
|     | DeepEns-Lite      | 0.9858 | 0.9812 | 0.9781  | 0.9774 | 0.9831 | 0.9803 |
|     | DeepEns           | 0.9860 | 0.9887 | 0.9874  | 0.9818 | 0.9846 | 0.9817 |

### 4.4 Statistical Significance Test

The comparison of DeepEns with existing approaches is performed in Table 7. To avoid the confusion of superiority of one approach with the other, we have performed statistical significance test with one variable t-test. The hypothesis testing is performed with 95% confidence interval with the null hypothesis stated as "the mean value of correlation coefficient of row algorithm is greater than the value of column algorithm. In Table 7, a value of '0' indicates an indistinguishable scenario whereas a '1' indicates that the row algorithm is superior to the column algorithm and '-1' otherwise.

**Table 7** One variable t-test to check statistical significance

|      | [14] | [12] | [16] | [17] | [18] | [19] | [20] | [21] | [36] | [36] | [37] | [38] | [39] | [40] | [41] | [42] | [43] | Our |
|------|------|------|------|------|------|------|------|------|------|------|------|------|------|------|------|------|------|-----|
| **[14]** | 0 | -1 | -1 | -1 | 1 | 0 | -1 | -1 | -1 | -1 | -1 | -1 | -1 | -1 | -1 | 1 | -1 | -1 |
| **[12]** | 1 | 0 | 1 | 1 | 1 | 1 | 1 | 1 | 1 | 1 | 1 | 1 | 1 | 1 | 1 | 1 | 1 | -1 |
| **[16]** | 1 | -1 | 0 | -1 | 1 | -1 | -1 | 1 | -1 | 1 | 1 | -1 | 1 | -1 | 1 | 1 | 1 | -1 |
| **[17]** | 1 | -1 | 1 | 0 | 1 | -1 | -1 | 1 | -1 | 1 | 1 | 1 | 1 | 1 | -1 | -1 | 1 | -1 |
| **[18]** | -1 | -1 | -1 | -1 | 0 | 1 | -1 | 1 | -1 | -1 | -1 | -1 | -1 | -1 | -1 | -1 | 1 | -1 |
| **[19]** | 0 | -1 | 1 | 1 | -1 | 0 | -1 | 1 | 1 | 1 | 0 | 0 | 0 | 0 | 1 | -1 | 1 | -1 |
| **[20]** | 1 | -1 | 1 | 1 | 1 | 1 | 0 | 1 | 1 | 1 | 1 | -1 | 1 | 1 | 1 | 1 | 1 | -1 |
| **[21]** | 1 | -1 | -1 | -1 | -1 | -1 | -1 | 0 | -1 | -1 | 1 | -1 | -1 | -1 | -1 | -1 | 1 | -1 |
| **[36]** | 1 | -1 | 1 | 1 | 1 | -1 | -1 | 1 | 0 | 1 | 1 | -1 | 0 | -1 | -1 | -1 | 1 | -1 |
| **[36]** | 1 | -1 | -1 | -1 | 1 | -1 | -1 | 1 | -1 | 0 | 1 | -1 | -1 | -1 | -1 | -1 | 1 | -1 |
| **[37]** | -1 | -1 | -1 | -1 | 1 | 0 | -1 | -1 | -1 | -1 | 0 | -1 | -1 | -1 | -1 | -1 | -1 | -1 |
| **[38]** | 1 | -1 | 1 | -1 | 1 | 0 | 1 | 1 | 1 | 1 | 1 | 0 | 1 | -1 | 1 | 1 | 1 | -1 |
| **[39]** | 1 | -1 | -1 | -1 | 1 | 0 | -1 | 1 | 0 | 1 | 1 | -1 | 0 | -1 | 1 | 1 | 1 | -1 |
| **[40]** | 1 | -1 | 1 | -1 | 1 | 0 | -1 | 1 | 1 | 1 | 1 | 1 | 1 | 0 | 1 | 1 | 1 | -1 |
| **[41]** | 1 | -1 | -1 | 1 | 1 | -1 | -1 | 1 | 1 | 1 | 1 | -1 | -1 | -1 | 0 | -1 | 1 | -1 |
| **[42]** | -1 | -1 | -1 | 1 | 1 | 1 | -1 | 1 | 1 | 1 | 1 | -1 | -1 | -1 | 1 | 0 | 1 | -1 |
| **[43]** | 1 | -1 | -1 | -1 | -1 | -1 | -1 | -1 | -1 | -1 | 1 | -1 | -1 | -1 | -1 | -1 | 0 | -1 |
| **Our** | 1 | 1 | 1 | 1 | 1 | 1 | 1 | 1 | 1 | 1 | 1 | 1 | 1 | 1 | 1 | 1 | 1 | 0 |

**Table 8** Training on LIVE, CSIQ and TID2013 and testing on natural distortion databases

| Dataset → | CID2013 | | Live CD | | KonIQ-10K | | BIQ2021 | |
|---|---|---|---|---|---|---|---|---|
| Metric → | PLCC | SROCC | PLCC | PLCC | PLCC | SROCC | PLCC | SROCC |
| DeepEns | 0.6977 | 0.6762 | 0.6054 | 0.6878 | 0.6862 | 0.6917 | 0.6878 | 0.6724 |

### 4.5 Many-vs-One Cross-Dataset Evaluation

Evaluation of an approach through cross-validation is a good measure of its performance. Image quality assessment databases are relatively small-sized, have followed constrained subjective scoring experiments and therefore exhibit a lot of variation. Moreover, the databases with simulated distortion such as LIVE, TID2013 and CSIQ contain specific distortion categories with a discrete level of degradation on a set of reference images and therefore are not the ideal candidate for evaluation of image quality assessment. We have therefore proposed to train the IQA algorithm on simulated distortion databases (i.e. LIVE, TID2013, CSIQ) and cross-validate them on image databases with natural distortion. It is to be noted that different image databases use different scales for subjective scoring, therefore, we have normalized the subjective score of all the training and validation databases from 0-1. We have three image databases with natural distortions two are benchmark databases and one is self-collected image database. The results are reported in Table 8.

### 4.6 Experiments on BIQ2021

We have conducted three experiments for BIQ2021 image quality assessment database. In the first experiment, the synthetic distortion databases (LIVE, TID2013 and CSIQ) are used for training and the 2000 test images of BIQ2021 are used for testing. In the second experiment, naturally distorted databases (LIVE CD, CID2013 and KonIQ-10k) are used for training and BIQ2021 test-set of 2000 images is used for testing. Whereas the third experiment both synthetic and naturally distorted databases LIVE, CSIQ, TID2013, LIVE CD, CID2013, KonIQ-10K and 10,000 images of BIQ2021 are used for training and 2000 images of BIQ2021 are used for testing. The fourth experiment uses 10,000 images of BIQ2021 for training and 2000 images of BIQ2021 for validation. The results of these four experiments are reported in Table 9.

**Table 9** Cross-Dataset testing on BIQ2021

| Sr. | Train Set | Metric | BIQ2021 (Test Set) |
|---|---|---|---|
| 1 | Synthetic distortion databases | PLCC | 0.6878 |
|   |   | SROCC | 0.6924 |
| 2 | Natural distortion databases | PLCC | 0.8118 |
|   |   | SROCC | 0.8024 |
| 3 | Synthetic distortion + natural distortion databases | PLCC | 0.7427 |
|   |   | SROCC | 0.7351 |
| 4 | Train set of BIQ2021 | PLCC | 0.8098 |
|   |   | SROCC | 0.7922 |

**A. Ablation Study**

The ablation experiments are extensively incorporated in neuroscience to check these complex systems. In the area of Artificial Neural Networks (ANN), these experiments are done to check if all the parts of the architecture are necessary for the required performance. As the ANN are complex architectures and they don't have a set of rules for their construction, rather they follow heuristics for their construction. It is, therefore, necessary to verify if the designed architecture as a whole is necessary or some part of the architecture can be removed with no decrease in the intended performance of the ANN. We have done the experimentations during the construction of the architecture with low epochs to select the most suitable architecture for the problem at hand, however, the final architecture is pruned at some bottleneck points to verify the efficacy of its key components. We have used CSIQ database for ablation study and tested three different subsets of the architecture for validation. The pruned architecture is retrained for 100 epochs with CSIQ database and the experiment is repeated 10 times to report the median values of PLCC and SROCC provided in Table 10.

In the first experiment, we have pruned the Vgg16 subset of the architecture and verified the model with agreed parameters. In the second experiment, we have pruned the AlexNet subset and performed the validation. In the last experiment, the flatten layer is directly connected to the regression output layer. Table 10 can be viewed for the performance of these three ablation experiments.

**Table 10** Results of Ablation Study

| Sr. | Ablation Experiment | RMSE | PLCC | SROCC |
|---|---|---|---|---|
| 1 | Pruning EfficientNet-base | 0.1419 | 0.8473 | 0.8176 |
| 2 | Pruning NasNet-base | 0.1628 | 0.8688 | 0.8294 |
| 3 | Pruning Fully-connected layers | 0.1383 | 0.9233 | 0.8559 |

**B. Time Complexity**

The experimental setup is described at the start of this section. The training of DeepEns for fine-tuning on image quality database BIQ2021 is performed for 100 epochs for a batch size of 16 and it took almost 168 hours. Whereas in the testing phase, it takes 2.77 seconds per image. The testing is reported based on CPU only.

## 5 Discussion

The image quality assessment is a challenging task due to its relative nature. The perceptual quality of an image depends on several parameters and is largely influenced by the nature of content in the image. A large number of researchers have proposed different statistical and other features which are affected by a change in image quality but they are unable to fully cover the complexities of the factors affecting the image quality. Natural scene statistics-based quality assessment algorithms designed for a certain distortion type perform well only to those distortions and doesn't generalize to scenarios containing other types of distortions or combination of distortions. Some datasets are designed with multiple distortions [44, 45] but they again have the same issue that the training algorithm learns specific distortions only. In case of images with natural distortions such as [30], the algorithms trained on synthetic distortions doesn't perform well on these datasets as demonstrated in section 4.5 and 4.7. Training and validation on the naturally occurring distortions may perform better but they are not true representative of no-reference quality assessment scenario in general.

The proposed approach (DeepEns) has targeted the problem by training and testing it on five benchmark datasets. The results of individual dataset scenarios are reported and comparison of synthetic distortion databases is made with the existing approaches. The comparison is made based on correlation coefficients. DeepBIQ [12] is the second-best performing model which has used Caffe [46] architecture which is inspired from AlexNet as learning architecture and fine-tuned it on the dataset under test. Their architecture has performed exceptionally well on several benchmarking datasets.

BLINDER [16] has used Vgg to extract deep activations at multiple levels and trained an ensemble to make prediction and has reported second-highest scores for CSIQ dataset and comparable scores for other datasets. DIQA [17] have used a two-stage approach and reported second-highest scores for LIVE dataset and comparable scores for other datasets. One variable t-test is performed to check the statistical significance and the DeepEns has provided superior performance. The DeepEns has used CNN base of two architectures performed feature pooling which is passed through few fully connected layers with dropout. The proposed DeepEns architecture is retrained on the dataset under test until convergence and then validated. This deep ensembling strategy has provided a set of representations with diverse features. This architecture in result provides a richer feature set that can learn quality aware features. The pretraining of the proposed architecture on the synthetic database and then retraining on BIQ2021 with natural distortion make it a high performing model.

The cross-dataset experiment is conducted to train the model on synthetic distortion databases and validate them on three natural distortion databases to check for generalization. The results of this experiment are encouraging and indicate good generalization. We again credit the better generalization to the ensembling behavior of the DeepEns architecture and its retraining providing quality-aware representations. The cross-dataset comparisons are not made as DIQA [17] has performed a cross-dataset experiment by taking a subset of four distortion types only from CSIQ and TID2013 dataset as LIVE dataset contain only these four type of distortions. It is to highlight that the purpose of cross-dataset experiment is to check the generalizability of the model on other distortion types and scoring methods. Some of the approaches such as DeepBIQ [12] and BLINDER [16] have not reported such results altogether.

Experiments on BIQ202 are conducted to further validate the generalization of the proposed model. Some interesting observations in these experiments are:

a) DeepEns trained on synthetic distortion databases generalize very poorly on BIQ2021 test-set as the training dataset contains a fixed number of simulated distortions which are not representative of natural distortions.
b) DeepEns trained on natural distortion databases performed well on the test-set of BIQ2021 as the distortions available in the training set are closer to the distortions occurring in the test set.
c) The experiment based on combination of synthetic distortion as well as natural distortion database performed better than synthetic distortion database alone but worse than the natural distortion database.
d) The last experiment, where the train set of BIQ2021 is used for training and the test set is used for validation has provided the highest performance which is slightly superior to natural distortion databases. The reason could be that train and test set of this database are scored by same peoples in similar type of subjective scoring experiments.

It is therefore evident from these experiments that synthetic distortion database is not representative of naturally occurring image distortions. Synthetic distortion databases, although widely available can be used to assess image quality in scenarios where similar types of distortions are occurring such as assessment of image quality of compression algorithms. Still, this type of comparison will be less suitable as the model trained on compressed images will be biased towards existing compression schemes used in mode training. Whereas the natural distortion database having distortions added during the process of image acquisition, storage and post-processing are better representatives. Distortions in these images occur due to hardware limitation, compression for storage purposes or post-processing for enhancement, cropping or adjustment. Therefore, large-scale image databases such as BIQ2021 are a better candidate to train or judge the performance of a general-purpose image quality assessment model.

# 6 Conclusion and Future Work

In this study, we have explored the problem of general-purpose image quality assessment. We have proposed a deep ensemble-based architecture DeepEns which is first trained on synthetic database of 18.75 million images and then retrained on natural distortion database of 12,000 subjectively scored images. The proposed approach has performed consistently better in comparison to the existing approaches in single database experiment. In cross-dataset experiment, we have trained the model on synthetic databases and validated it on two natural distortion benchmark databases and one self-collected natural distortion database. The results of these experiments indicate that the DeepEns is robust and performs well when trained on a representative image database. To make a general-purpose image quality assessment algorithm, the DeepEns is trained on three natural distortion databases and validated on an independent subset of natural distortion databases. The final trained model (DeepEns) and the self-collected database (BIQ2021) will be made publicly available for comparison and benchmarking.

## 6.1 Future Work

In the future work we will address the following directions:
a) Improvement in the generalization of the trained model by training on a combination of different natural distortion databases after realignment and normalization of their subjective scores.
b) Making specific design decisions to further improve the generalization performance of end-to-end trained CNN architecture.
c) Use of quality aware or visual perception-based loss function for model training.

# APPENDIX A

**Loss Functions**

Choice of the loss function is important while training a regression algorithm. IQA is a regression problem, therefore careful selection of suitable loss function is important. We have experimented with six loss functions. Although, the loss function can be chosen intuitively among the candidates but experimenting with different loss functions is a better way. The details of the six loss functions and the training progress is provided from Fig. X-X. The training is performed by using the LiveCD database and trained for 30 epochs.

**Mean Squared Error (MSE)**

MSE is the most commonly used loss function for regression. It is sum of the squared distances between the target and predicted values and is calculated by:

$$MSE = \frac{1}{n}\sum_{n}(T - P)^2$$

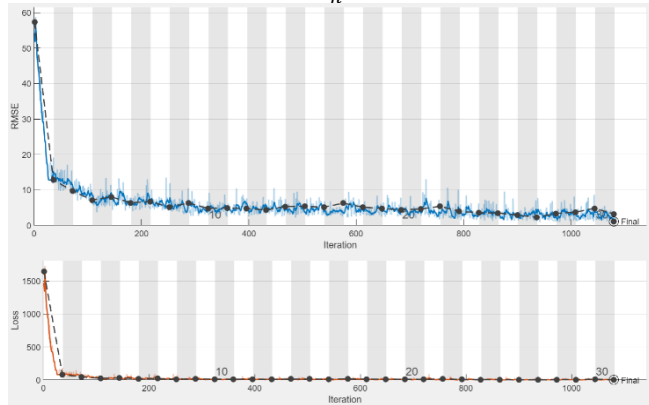

**Fig. A-1** Training progress using MSE as loss function

**Mean Absolute Error (MAE)**
MAE is another useful loss function and is robust to outliers. It measures the directionless magnitude of the errors by taking sum of absolute differences between target and predictions. It is calculated by the following formula:

$$MAE = \frac{1}{n}\sum_{n}|T - P|$$

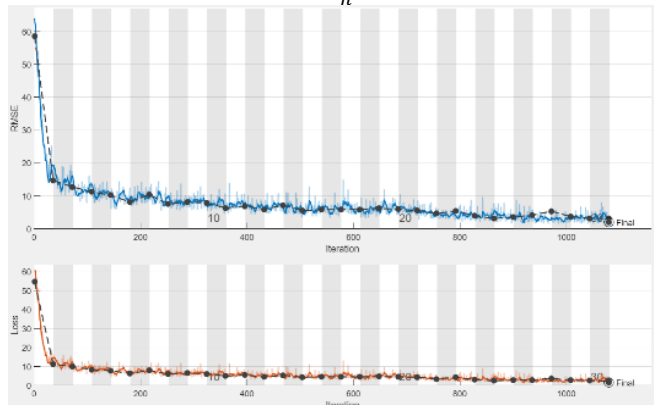

**Fig. A-2** Training progress using MAE as loss function

**Mean Absolute Percentage Error (MAPE)**
MAPE provides the percentage magnitude of the error which is calculated by dividing the error calculated by MAE by the target value. Although it is simple and convincing, it has the drawback of not being useful when there are target values of zero. Moreover, it puts havier penalty on negative errors where the forecasted value is higher than the actual value and therefore it provides outcomes that are less than the target value. The formula for calculation of MAPE is provided below:

$$MAPE = \frac{1}{n}\sum_{n}\left|\frac{T-P}{T}\right|$$

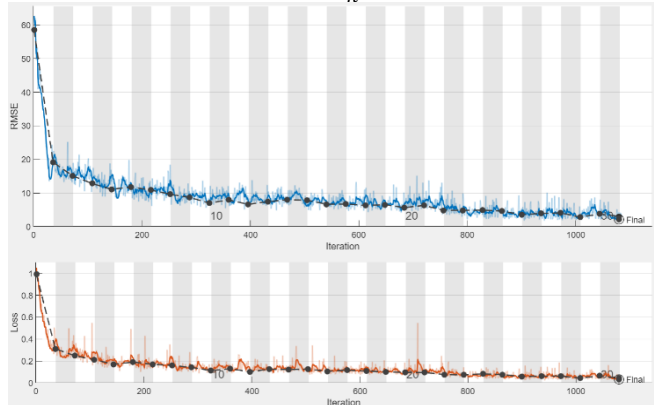

**Fig. A-3** Training progress using MAPE as loss function

**Mean Squared Logarithmic Error (MSLE)**

MSLE is a modified version of MSE and acts as a measure of ratio between the target and the predicted values. As it is a ratio and cares more about the percentage difference between the target and predicted values so it treats large and small differences similarly. Moreover, it is also asymmetric like MAPE but it favors larger predictions more than smaller predictions. The following formula is used to calculate MSLE:

$$MSLE = \frac{1}{n}\sum_{n}(\log(T+1) - \log(P+1))^2$$

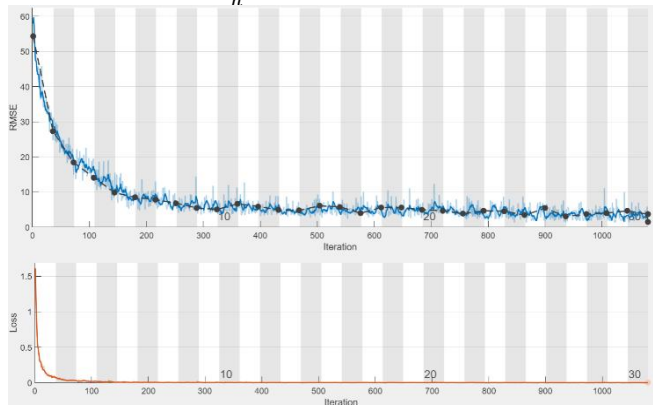

**Fig. A-4** Training progress using MSLE as loss function

**Huber Loss**

It is an alternative loss function to MSE and MAE which combines the good properties of both of them and is differentiable at 0. For small prediction error, it acts like MSE and for larger prediction error it acts like MAE and therefore is robust to outliers and provides better convergence when the loss is near minima. The drawback with using hubber loss is that we have to tune the hyperparameter $\delta$ as its value will define the choice of piecewise function. The hubber loss can be calculated using the following formula, it is to be noted that we have not tuned it for hyperparameter $\delta$ and used '1' as its value:

$$Huber = \begin{cases} \frac{1}{2}(T-P)^2 & for\ (T-P) \leq \delta \\ \delta|T-P| - \frac{1}{2}\delta^2 & otherwise \end{cases}$$

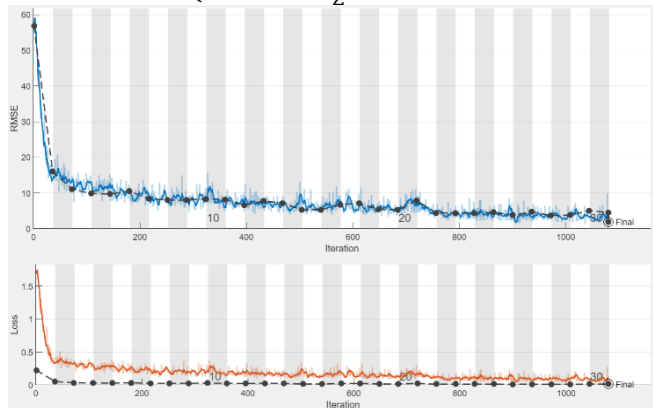

**Fig. A-5** Training progress using Hubber loss

**LogCosh Loss**
It measures the logarithm of the hyperbolic cosine of the prediction error and is smoother than MSE. It acts like MSE for smaller prediction error and is not strongly affected by an intermittent large prediction error and is therefore advantageous to Hubber loss. The formula for calculation of LogCosh loss is provided below:

$$LogCosh = \sum_n \log(\cosh(P-T))$$

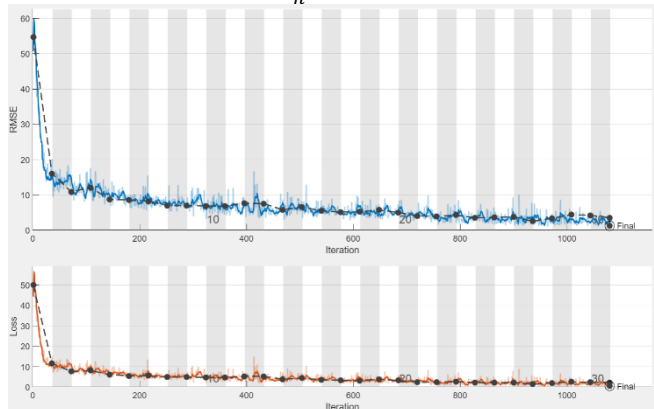

**Fig. A-6** Training progress using LogCosh loss

## Declarations

**Funding:** Not applicable.
**Conflicts of Interest:** The authors declare that there is no conflict of interest.
**Availability of data and material:** The data is currently available in GitHub repository: https://github.com/nisarahmedrana/DeepEns
**Code availability:** https://github.com/nisarahmedrana/DeepEns
**Authors' contributions:** All authors contributed equally.

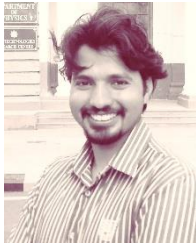**Nisar Ahmed** is Ph.D. student in the Department of Computer Engineering, University of Engineering and Technology, Lahore, Pakistan. His areas of interest include Image Quality Assessment and Computer Vision.

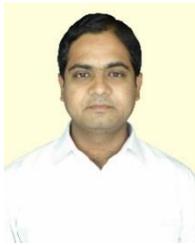**Hafiz Muhammad Shahzad Asif** obtained his Ph.D. degree in informatics from the University of Edinburgh, UK in 2012. He is working as Chairman and Associate Professor at the Department of Computer Science, University of Engineering and Technology, Lahore (New Campus), Pakistan.